\documentclass[10pt,twocolumn,letterpaper]{article}

\usepackage{cvpr}
\usepackage{graphicx}
\usepackage{comment}
\usepackage{amsmath,amssymb} 
\usepackage{color}
\usepackage{times}
\usepackage{epsfig}

\usepackage{amsfonts}       
\usepackage{amsthm}   

\DeclareMathOperator*{\argmin}{arg\,min}
\usepackage{bm}
\usepackage{algorithm}
\usepackage{algorithmic}

\newcommand{\tabincell}[2]{\begin{tabular}{@{}#1@{}}#2\end{tabular}}

\newcommand{\btheta}{{\bm{\theta}}}
\newcommand{\bT}{{{\bf T}}}
\newcommand{\N}{{\mathcal{N}}}

\newcommand{\bI}{{\bf I}}

\newcommand{\be}{\begin{eqnarray}}
\newcommand{\ee}{\end{eqnarray}}
\newcommand{\beq}{\begin{equation}\begin{aligned}}
\newcommand{\eeq}{\end{aligned}\end{equation}}
\newcommand{\beqn}{\begin{equation*}\begin{aligned}}
\newcommand{\eeqn}{\end{aligned}\end{equation*}}
\newcommand{\ben}{\begin{eqnarray*}}
\newcommand{\een}{\end{eqnarray*}}

\usepackage{algorithm}
\usepackage{algorithmic}

\usepackage{multirow}
\usepackage{booktabs}       

\usepackage[pagebackref=true,breaklinks=true,letterpaper=true,colorlinks,bookmarks=false]{hyperref}

\usepackage{authblk}
\cvprfinalcopy
\ifcvprfinal\fi
\begin{document}

\title{Stochastic Model Pruning \\ via Weight Dropping Away and Back} 
\author{Haipeng Jia}
\author{Xueshuang Xiang\thanks{Corresponding author: xiangxueshuang@qxslab.cn}}
\author{Dan Fan}
\author{Meiyu Huang}
\author{Changhao Sun}
\author{Yang He}
\affil{\normalsize Qian Xuesen Laboratory of Space Technology \\ China Academy of Space Technology}
\date{}

\maketitle

\begin{abstract}
  Deep neural networks have dramatically achieved great success on a variety of challenging tasks. However, most successful DNNs have an extremely complex structure, 
leading to extensive research on model compression.
As a significant area of progress in model compression, traditional gradual pruning approaches involve an iterative prune-retrain procedure and may suffer from two critical issues: \emph{local importance judgment}, where the pruned weights are merely unimportant in the current model; and an \emph{irretrievable pruning process}, where the pruned weights have no chance to come back. 
Addressing these two issues, this paper proposes the Drop Pruning approach, which leverages stochastic optimization in the pruning process by introducing a drop strategy at each pruning step, namely, drop away, which stochastically deletes some unimportant weights, and drop back, which stochastically recovers some pruned weights. 
The suitable choice of drop probabilities decreases the model size during the pruning process and helps it flow to the target sparsity. 
Compared to the Bayesian approaches that stochastically train a compact model for pruning, we directly aim at stochastic gradual pruning. 
We provide a detailed analysis showing that the drop away and drop back approaches have individual contributions. 
Moreover, Drop Pruning can achieve competitive compression performance and accuracy on many benchmark tasks compared with state-of-the-art weights pruning and Bayesian training approaches. 
\end{abstract}

\section{Introduction}
In recent years, various kinds of deep neural networks (DNNs) have dramatically improved the accuracy in many artificial intelligence tasks, 
including computer vision tasks, from basic image classification challenges~\cite{alexnet,vgg2014,resnet} to some advanced applications, e.g., object detection~\cite{liu2016ssd} and semantic segmentation~\cite{noh2015learning}. 
However, these networks generally contain tens of millions of parameters, leading to considerable storage requirements, 
which increase the difficulty of applying DNNs on the devices with limited memory~\cite{cheng2017survey}. 

\begin{figure*}[tb]
	\begin{center}
		\includegraphics[width=1.0\textwidth]{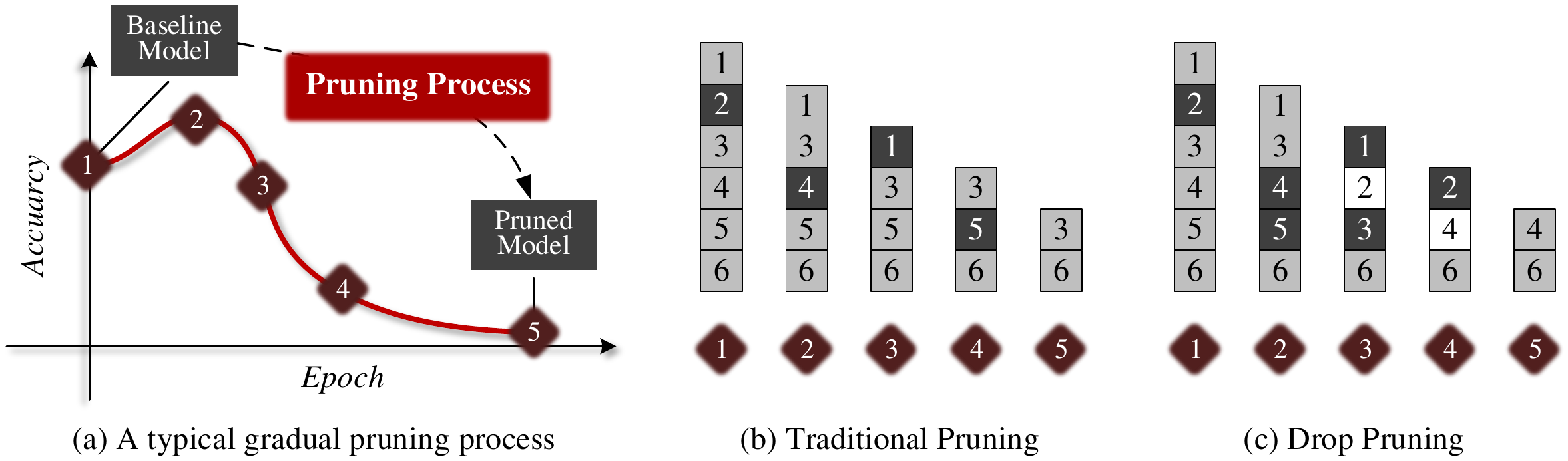}
	\end{center}
	\caption{(a): An illustration of the model accuracy varying in a typical gradual pruning process, (a)$\&$(b): the Traditional Pruning process, and (a)$\&$(c): the proposed Drop Pruning. The vector with entries from $1$ to $6$ denotes the baseline model with 6 weights. Traditional pruning processes iteratively delete some unimportant weights and then retrain the model, in which \emph{the current pruned model is a subset of the previous one}. In contrast, Drop Pruning iteratively drops away some unimportant weights, then drops back some pruned weights and retrains, in which \emph{the current pruned model is smaller than the previous one}. We use the white character on black background to represent the weights that will be dropped away at the next pruning step and the black character on white background to represent the weights that are dropped back at this pruning step. As shown in (c), at the second pruning step, Drop Pruning drops away the weights $\{4,5\}$ and drops back the weight $\{2\}$. Finally, Drop Pruning outputs the pruned model with $2$ weights but of different locations compared with the Traditional Pruning. }\label{fig1}
\end{figure*}

One way to address the above issue is model compression, since models are always greatly overparameterized~\cite{ullrich2017soft,molchanov2017variational}. 
Among varying model compression approaches, model pruning has made significant progress. 
As shown in Figure \ref{fig1} (a)$\&$(b), starting from a baseline model (the uncompressed model, denoted by a vector), a Traditional Pruning process~\cite{han2015learning,zhu2017prune} first deleted some ``unimportant'' weights (the entries in the vector) and then retrained the model. After deleting and retraining several times, the pruning process outputs a pruned model (a smaller vector). However, we attack the Traditional Pruning by addressing the following two critical issues: 
\begin{itemize}
	\item {\bf The importance judgment is local.} A normal way is determining the weights' importance at each pruning step, for instance, by their magnitude~\cite{han2015learning}. However, since the interconnections among the weights are so complicated, the weights' importance may change dramatically during pruning, i.e., the importance is just a local judgment, which means that \emph{the weights of less importance at this time may become more important in the future}. 
	\item {\bf Once pruned, there is no chance to come back}. If we view the pruning problem as an $\mathcal{L}_0$ optimization problem of weights~\cite{louizos2017learning}, the pruning process in Figure \ref{fig1} (a)$\&$(b) will shrink the optimization domain (the pruned weights have no chance to come back); thus, the optimization process \emph{has no chance to escape from the local minimum\footnote{This local minimum refers to the pruning strategy, i.e., the one that minimizes the $\mathcal{L}_0$ term. Another local minimum that is always mentioned in deep learning refers to the one involved with learning, i.e., it minimizes the loss term of the data $\mathcal{L}_D$.}}. 
\end{itemize}

Considering these issues, this paper leverages the stochastic optimization technique in the pruning process. 
At each pruning step, we first delete the unimportant weights with some probability, named drop away, which help it 
avoid deleting global important weights. 
Then, we also recover some weights from the deleted weights with some probability, named drop back, which help it 
reload the pruned global important weights. We named our approach Drop Pruning since the crucial step in optimizing the pruning process is the drop strategy. 
For instance, as shown in Figure \ref{fig1} (a)$\&$(c), at the second pruning step, we drop away the weights $\{4,5\}$ and drop back the weight $\{2\}$, at the third pruning step, we drop away the weights $\{1,3\}$ and drop in back the weight $\{4\}$. 
At last, Drop Pruning also outputs the pruned model with only $2$ weights but of different locations compared with the Traditional Pruning process in Figure \ref{fig1} (a)$\&$(b). 
Unlike Bayesian training approaches~\cite{kingma2015variational,lee2018adaptive,dai2018compressing,Adian2018Targeted} that impose stochastic training on a compact model for pruning, Drop Pruning directly imposes stochastic pruning that may help the model flow to a simpler network. 

We demonstrate the performance of Drop Pruning on LeNet and VGG-like architectures using the MNIST and CIFAR-10 datasets. 
Our experiments show that drop away and drop back have individual contributions compared with the Traditional Pruning. 
Moreover, Drop Pruning can also achieve competitive compression and accuracy performance, compared with the related weights pruning and Bayesian training approaches. 
The proposed method may provide new insights in the aspect of model compression. 

\section{Drop Pruning}

\subsection{Notations}
Denote $\bm{\theta}$ a DNN model and $\bT$ a pruning state, a binary vector with the same size of $\btheta$. Denote $(\btheta, \bT)$ a pruned model, in which the entries of $\bT$ indicate the state of model $\btheta$ ($0$ means the corresponding entry in $\btheta$ has been pruned).  Denote $\bI$ the ones vector of the same size as $\btheta$. Denote $[\mathbf{x}]$ the dimensionality of a vector $\mathbf{x}$. Suppose $\N(\bT)$ is a set that contains the locations of ones in a binary vector $\bT$, i.e., the indexes of unpruned weights. 
Denote $\mathcal{L}_D(\btheta)$ the loss function on the data. 

\subsection{Perspective of $\mathcal{L}_0$ Regularization}
$\mathcal{L}_0$ regularization of (blocks of) the weights of a model is a conceptually attractive approach of model compression 
and is always formulated as the following optimization problem:
\beq\label{L0term}
\btheta^* = \argmin_{\btheta} \mathcal{L}_D(\btheta) +\lambda \|\btheta\|_0,
\eeq
where $\lambda$ is a weighting factor for regularization. 
Different choices of $\lambda$ will lead to varying model selection criteria, such as the Akaike information criterion~\cite{akaike1998information} 
and the Bayesian information criterion~\cite{schwarz1978estimating}. In the area of sparse optimization, given the loss term $\mathcal{L}_D(\cdot)$ 
is a convex operator about $\btheta$ (a penalty term, such as $\|\mathbf{A}\btheta - \mathbf{b}\|_2^2$ with matrix $\mathbf{A}$ and vector $\mathbf{b}$ as implemented in image restoration, sparse coding, compressed sensing or Lasso), a common approach for solving \eqref{L0term} is iterative shrinkage/thresholding (IST)~\cite{daubechies2004iterative}. For the $\mathcal{L}_0$ penalty\footnote{IST instead uses soft thresholding for $\mathcal{L}_1$ penalty.}, the IST iteration for the \eqref{L0term} serves as:
\beq\label{IST}
\btheta^{k+1} = \underbrace{\Psi_{{\rm Hard}, \lambda/\alpha}}_{\text{Pruning}}
\underbrace{\left ( \btheta^{k} - \frac{1}{\alpha} \cdot \frac{\partial \mathcal{L}_D}{ \partial \btheta^k} \right )}_{\text{Retrain}}, 
\eeq
where $1/\alpha$ is the learning rate and $\Psi_{{\rm Hard}, s}$ is the hard thresholding with threshold $s>0$, deleting the entries in $ \btheta$ with magnitude less than $s$. 
Taking a glance at the Traditional Pruning process in Figure \ref{fig1} (a)$\&$(b), we consider it as corresponding to an extension of IST (retrain several steps and prune), as marked below \eqref{IST}. However, we cannot expect its convergence to the global minimum since $\mathcal{L}_D(\cdot)$ is usually not convex in Deep Learning. From this perspective, the pruning-based model compression approaches~\cite{han2015learning,zhu2017prune} can be integrated into the framework of $\mathcal{L}_0$ regularization. 
Additionally, in this framework, the Bayesian training approaches~\cite{kingma2015variational} focus on the stochastic design of $\lambda$ (or the penalty on weights' prior), and someone directly considers approximating $\mathcal{L}_0$~\cite{louizos2017learning}. Our approach serves as a modification on the pruning operator, i.e., the thresholding operator for the sparse optimization problem with a nonconvex penalty term. 

\subsection{Weight Dropping Away}
A very straightforward approach to reduce model complexity is gradually pruning the ``unimportant'' weights by their magnitude~\cite{han2015learning,zhu2017prune} 
or Hessian information~\cite{lecun1990optimal,hassibi1993second}. However, this information can only reflect the weights' importance of the current model, 
not the desired optimal model of the underlying classification or regression problem. 
We attack this problem by introducing the stochastic approach into the pruning process, in which the intent is to optimize the weights' importance judgment. 
Our heuristic motivation is that the weights with a lower magnitude are not necessarily less important but that they may have a high probability to be less important. 

\begin{figure*}[tb]
	\centering
	\includegraphics[width=0.6\textwidth]{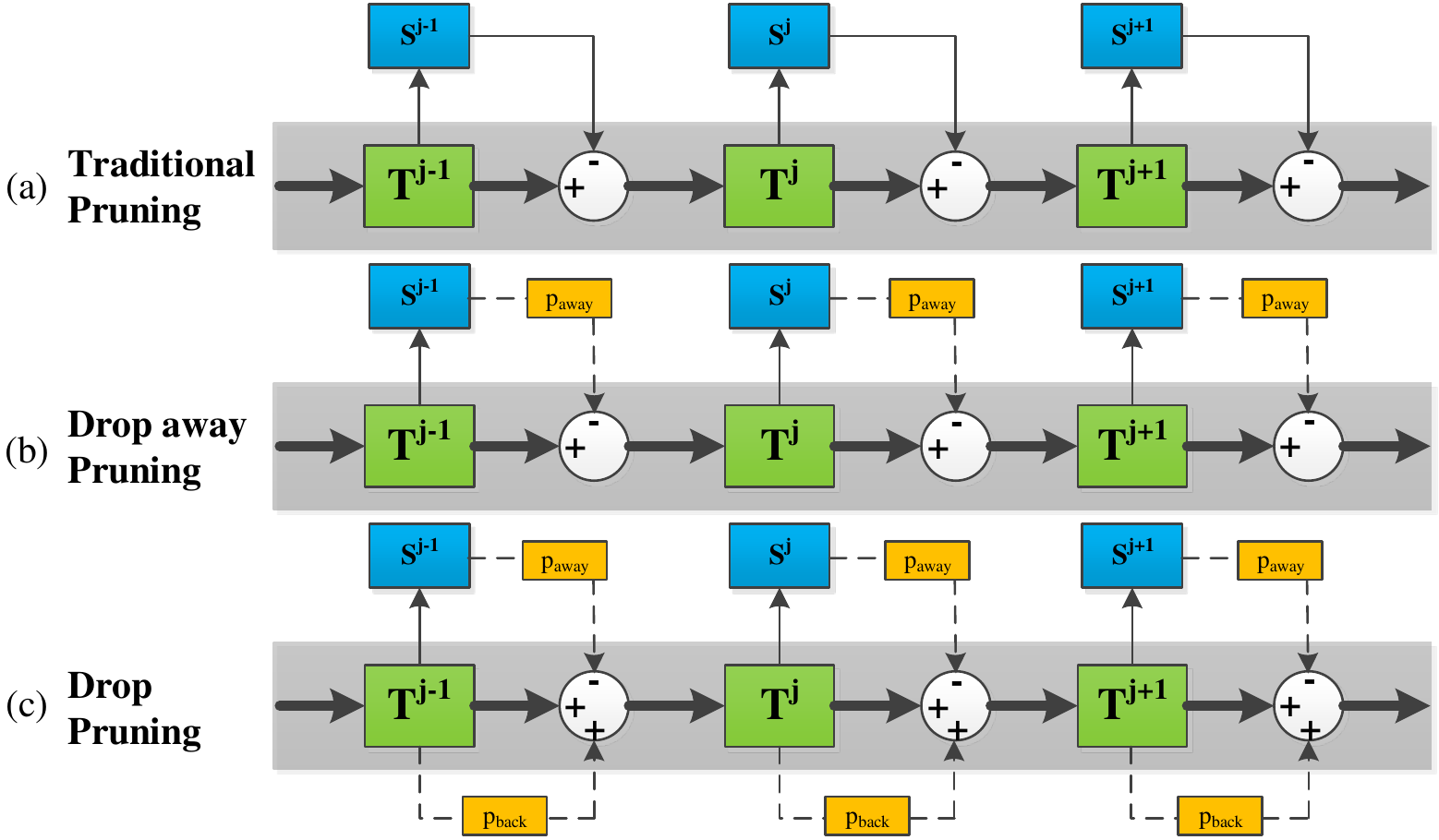}
	\caption{Pruning flow of (a) Traditional Pruning, (b) Drop away Pruning and (c) Drop Pruning. 
	}\label{fig4}
\end{figure*}

The traditional gradual pruning method~\cite{han2015learning,zhu2017prune} starts from the baseline model, i.e., setting $\bT^0:=\bI$; at each pruning step $j$, it firstly finds the ``unimportant'' ($|\btheta^j_i| \leq \lambda$) and unpruned ($\bT^{j}_i = 1$) weights of model $\btheta^j$, 
\beq\label{unimportant_weights}
\bm{S}^j = \{i\, | \,|\btheta^j_i| \leq \lambda, \bT^{j}_i = 1\}, 
\eeq
and then updates $\bT$ by $\bT^{j+1}_i = 0, \,{\rm if}\, i \in \bm{S}^j$, 
where $\lambda$ is a predefined threshold that can vary in different pruning steps and layers. 
We also denote $\bm{S}^j$ the set of pre-pruned weights. 
After deleting the pre-pruned weights, we retrain the pruned model (only retraining the unpruned weights). 
Then, $\bT $ will be updated as shown in Figure \ref{fig4} (a), and we can easily check that $\N(\bT^{j+1}) \subset \N(\bT^{j})$, i.e., the pruned model is a subset of the previous one.
Instead of pruning all the pre-pruned weights, drop away stochastically updating $\bT$ by 
\beq\label{drop_out}
\bT^{j+1}_i = 0, \,{\rm if}\, i \in \mathcal{B}(\bm{S}^j, p_{\rm away}), 
\eeq
where $\mathcal{B}(\mathbf{M}, p)$ is the set that randomly contains $p_{\rm away}[\mathbf{M}]$ elements in set $\mathbf{M}$. 
At this time, drop away will lead $\bT $ updates, as shown in Figure \ref{fig4} (b). 
However, we still have $\N(\bT^{j+1}) \subset \N(\bT^{j})$, as that for Traditional Pruning. 
We denote the pruning process that only employs drop away as Drop away Pruning. 

In practice, we first generate a vector (it has the same size as $\mathbf{M}$) of independent random variables with a uniform distribution in $[0,1]$ and then keep the indexes of the $p[\mathbf{M}]$ smallest entries. Then, the entries in $\mathbf{M}$ of these indexes are output. We note that the distribution the selected entries follow is not Bernoulli.

\begin{figure}[tb]
	\centering
	\includegraphics[width=0.45\textwidth]{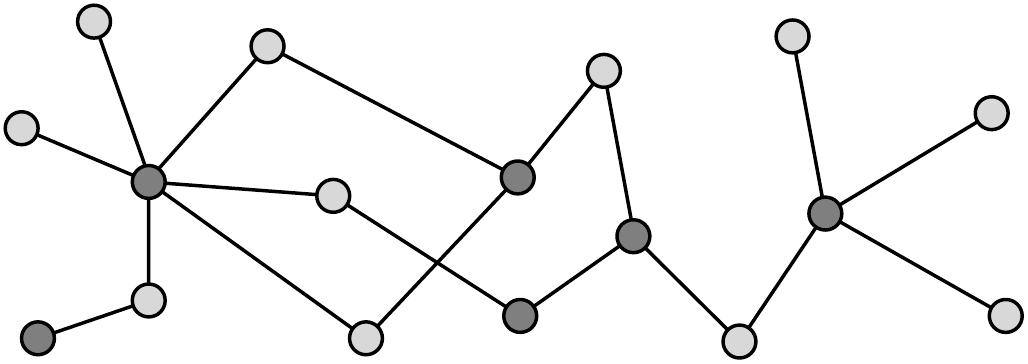}
	\caption{A minimal weighted vertex cover problem (MWVC): finding a minimal weighted set of vertices (solid ones) that touch all the edges in a given graph. 
		The drop back in Drop Pruning has similar spirit of the \emph{memory mechanism} in solving MWVC~\cite{Sun2018Potential}. }\label{fig2}
\end{figure}

\subsection{Weight Dropping Back}
Once the pruning process falls into a local minimum (the $\mathcal{L}_0$ optimization domain is shrunken, $\N(\bT^{j+1}) \subset \N(\bT^{j})$ both for Traditional Pruning and Drop away Pruning), it will have no chance to escape. 
Drop back is proposed to overcome this problem, and the main idea is randomly dropping back some weights from the pruned ones. 
Define the set $\bm{K}^j=\{i\, | \,\bT^{j}_i = 0\}$ that contains the indexes of pruned weights. 
After dropping away the unimportant weights by \eqref{drop_out}, we also drop back some pruned weights with probability $p_{\rm back}$: 
\beq\label{drop_in}
\bT^{j+1}_i &= 1, \,{\rm if}\,i \in \mathcal{B}(\bm{K}^j, p_{\rm back}), 
\eeq
which provides the model with a chance to escape from a local minimum of the $\mathcal{L}_0$ term. 
Furthermore, this action may also help the model to escape from a local minimum of the $\mathcal{L}_D$ term, 
similar to the Sparse-Dense action in DSD training~\cite{han2016dsd}. 
As shown Figure \ref{fig4} (c), since drop back will reload some pruned weights into the model, obviously, at this time, $\N(\bT^{j+1}) \nsubseteq \N(\bT^{j})$. 
In practice, we will suitably choose $p_{\rm away}$ and $p_{\rm back}$ to impose $[\N(\bT^{j+1})] \leq [\N(\bT^{j})]$, i.e., the pruned model is smaller than the previous one. 
Of course, it is also optional to set $p_{\rm away}$ and $p_{\rm back}$ as varying hyperparameters in different pruning steps and layers, similar to the threshold $\lambda$. 

{\bf Motivation from Integer Optimization}. 
Another motivation of drop back is derived from integer optimization. 
Integer and mixed integer constrained optimization problems~\cite{karlof2005integer} are NP-complete problems in optimization, in which some or all of the variables are restricted to be integers. 
As mentioned in~\cite{srivastava2014dropout}, a neural network with $n$ weights can be seen as a collection of $2^n$ possible thinned neural networks, such that searching the \emph{best} one is exactly an NP-hard integer optimization problem. The idea of the proposed drop back embodies a similar spirit to the \emph{memory mechanism} of an approach for solving the minimal weighted vertex cover (MWVC) problem~\cite{Sun2018Potential}. 

As shown in Figure \ref{fig2}, the well-known MWVC problem is as follows: given a graph $(\bm{V}, \bm{E})$, to find a minimal weighted set of vertices (a binary vector $\bm{S}$ with each entry denote whether we take the corresponding vertice or not), that touch all the edges in a given graph~\cite{taoka2012performance}. The MWVC problem can be formulated as: 
\beq \label{mwvc}
\bm{S}^*  = {\rm argmin}_{\bm{S}}\, \mathcal{F}_{\rm weight}(\bm{S};\bm{V},\bm{E}) + \|\bm{S}\|_0, 
\eeq
where $\mathcal{F}_{\rm weight}(\cdot)$ is the objective function that constrains the weight sum 
and the last term is the penalty on the sparsity of $\bm{S}$. 
In~\cite{Sun2018Potential}, the authors proposed a distributed algorithm for solving the MWVC problem, where each player (each vertex in the graph) simultaneously updates its action ($0$ or $1$) by obeying the relaxed greedy rule followed by a mutation with some probability, a mutative action is randomly drawn from the memory (the history actions). They found that if each player chooses the deterministic best response, the algorithm will converge to the local minimum that depends on the initial states. In contrast, if each player chooses a random action from its memory (\emph{memory mechanism}), the algorithm will converge to a better solution with high probability. 
The effectiveness and theoretical analysis of their proposed algorithm demonstrate that  stochastic optimization with the \emph{memory mechanism} is an effective technique in handling integer optimization problems.

For the model pruning problem formulated by the minimization problem with the $\mathcal{L}_0$ term, we can associate a correspondence of each weight to a vertex in a graph 
(regarding the similarity between \eqref{L0term} and \eqref{mwvc}). 
Thus, the drop back can be seen as a special kind of \emph{memory mechanism}, such that each weight has some probability of rebirth from its memory. 

\subsection{Implementation details}
$\quad$ {\bf Algorithm}. Combining the iterative scheme \eqref{IST} and drop away \eqref{drop_out} and drop back \eqref{drop_in}, the Drop Pruning algorithm is summarized in Algorithm \ref{alg1}. 
Given a baseline model and the target sparsity, the algorithm will output a satisfactory pruned model by iteratively performing the weight dropping and retrain steps. 
Here, the threshold $\lambda$ in each pruning step is computed by the target sparsity at the current pruning step, which is smoothly predefined with the last one as the final 
target sparsity. We refer to~\cite{zhu2017prune} for the details of the gradual pruning settings. 

{\bf Drop probabilities}. Let $\xi_1, \xi_2 \in [0,1]$. We choose the {drop away} probability $p_{\rm away}$ to be $\xi_1$. To make $[\N(\bT^{j})] \leq [\N(\bT^{j-1})]$, i.e., make the pruned model smaller than the last one, we set $p_{in} = \xi_2 [\bm{S}^{j-1}] / [\bm{K}^{j-1}]$ with $\xi_2 < \xi_1$. Then, each pruning step will result in 
$
[\N(\bT^{j})] = [\N(\bT^{j-1})] - p_{\rm away}[\bm{S}^{j-1}] + p_{\rm back}[\bm{K}^{j-1}] = [\N(\bT^{j-1})] - (\xi_1 - \xi_2)[\bm{S}^{j-1}]  < [\N(\bT^{j-1})]. 
$

\begin{algorithm}[tb]
	\caption{Drop Pruning algorithm for stochastic model compression. 
		Given a baseline model and the target sparsity, the algorithm will output a satisfactory pruned model by iteratively performing the weight dropping and retrain steps. 
		The threshold $\lambda$ of each pruning step is predefined by smoothing the target sparsity~\cite{zhu2017prune}. 
		$1 - [\N(\bT)] / [\bT]$ represents the sparsity of the pruned model $(\btheta,\bT)$. $\mathcal{B}(\bm{M}, p)$ represents the set which randomly contains $p[\mathbf{M}]$ entries in $\mathbf{M}$. Note the distribution of $\mathcal{B}(\bm{M}, p)$ is not Bernoulli. }
	\label{alg1}
	\begin{algorithmic}[1]
		\REQUIRE Baseline model $\btheta$, target sparsity $s$, drop probabilities $p_{\rm away}$, $p_{\rm back}$, learning rate $\gamma$; 
		\ENSURE Pruned model $(\btheta, \bT)$. 
		\STATE $\bT \gets  \bI$ \COMMENT{Define initial pruning state}		
		\WHILE{$(1 - [\N(\bT)] / [\bT]) \geq s$} 
		\STATE $\bm{S} \gets \{i\, | \,|\btheta_i| \leq \lambda, \bT_i = 1\}$; \COMMENT{Find the pre-pruned weights}
		\STATE $\bm{K} \gets \{i\, | \,\bT_i = 0\}$; \COMMENT{Find the pruned weights}
		\STATE $\bT_i \gets 0, \,{\rm if}\, i \in \mathcal{B}(\bm{S}, p_{\rm away})$; \COMMENT{Drop away some pre-pruned weights}
		\STATE $\bT_i \gets 1, \,{\rm if}\,i \in \mathcal{B}(\bm{K}, p_{\rm back})$; \COMMENT{Drop back some pruned weights}
		\FOR{several minibatches}
		\STATE $\btheta \gets \btheta - \gamma \cdot \partial \mathcal{L}_D / \partial (\btheta \cdot \bT)$; \COMMENT{Retrain}
		\ENDFOR
		\ENDWHILE
	\end{algorithmic}
\end{algorithm}

\section{Related work}
Model compression and acceleration is a popular research area, and tremendous progress has been made in the past few years; see the review papers~\cite{cheng2017survey,sze2017efficient} and the references therein for details. 
The most closely related category to our approach is model pruning, which involves gradually finds and deletes unimportant weights. 
Most existing approaches aim at designing the criterion for deciding importance, like magnitude-based approaches~\cite{han2015learning,zhu2017prune,lee2018snip,yeh2018deep}, which 
assume smaller weights have less importance but with no theoretical evidence, and Hessian-based approaches~\cite{lecun1990optimal,hassibi1993second,dong2017learning,zeng2018mlprune}, which 
can not be adopted to large neural network due to the large size and cost of obtaining Hessian matrix. 
All these methods may suffer from the critical issues of local importance judgment and irretrievable pruning process. 
Recently, attacking the later issue, Guo \textit{et al.}~\cite{dns} proposed dynamic network surgery (DNS), which incorporates weight splicing into the whole pruning process. However after each pruning step, DNS reload the whole network and the splicing is deterministic. 
In our work, the drop back has the same motivation with the splicing but instead using stochastic criterion.

Another related category is training compact model with Bayesian approaches, since we both introduce stochastic for obtaining a sparse model. 
The most modern approaches contain dropout~\cite{hinton2012improving,srivastava2014dropout,bouthillier2015dropout}, variational dropout~\cite{kingma2015variational}, targeted dropout~\cite{Adian2018Targeted} and its variants~\cite{neklyudov2017structured,lee2018adaptive,dai2018compressing,srinivas2016generalized,gal2017concrete}. However, our approach survives with a main observation: Bayesian approaches impose stochastic training of a compact model for pruning, whereas Drop Pruning instead directly imposes stochastic into pruning. 
Technically, if we set $p_{\rm back}=1$ and then recover all the pruned weights in drop back, Drop Pruning resembles targeted dropout, but with a different drop strategy: 
targeted dropout uses a Bernoulli distribution, i.e., each unit has a probability to be deleted or not at a particular pruning step, whereas Drop Pruning imposes that we should randomly prune at least a fixed number of units (with probability $p_{\rm away}$), i.e., the distribution is not Bernoulli. 
An interesting investigation is learning a compact model by targeted dropout and then pruning it by Drop Pruning. 

\section{Experiments}
In this section, we evaluate our Drop Pruning on various model compression benchmarks (LeNet-5 and LeNet-300-100 (LeNet-FC) on MNIST,  and VGG-16 on CIFAR10) to demonstrate its performance against 
the Traditional Pruning, the related weights pruning and Bayesian training approaches. Our results show that drop away and drop back have their individual contributions and Drop Pruning yields competitive compression and accuracy performance against the comparison approaches. All the experiments were implemented on a GPU cluster with 16 NVIDIA V100 GPUs (16 GB). Reproducing these experiments requires approximately 1 GPU year (NVIDIA V100). 

{\bf Benchmarks}. We consider three benchmarks: LeNet-5 and LeNet-300-100 on  MNIST~\cite{lecun1998gradient},  and VGG-16 on CIFAR10~\cite{simonyan2014very}. 
LeNet-5 is a convolutional network that has two convolutional layers and two fully connected layers. LeNet-300-100 is a feedforward neural network with three fully connected layers. VGG-16 has several convolutional layers but only three fully connected layers. In pruning process, the batch size, learning rate and learning policy are set as the same in training the baseline models. 

{\bf Off-line pruning}. Drop Pruning is a stochastic pruning strategy, and each trial will lead to a different pruned model. The following results are the ones under $40$ trials for LeNet-5 and LeNet-FC and $10$ trials for VGG-16, considering that VGG-16 has much cost. We also use the same strategy (training $40/10$ trials) for Traditional Pruning to take into account the influence of SGD-based training. 
Here we use the best one to represent the ability of the proposed algorithm, since the pruning can be done off-line. Once we obtain a pruned model off-line, the on-line inference is deterministic. 
Unlike the related weight pruning approaches which need specifying the target sparsity for each layer, we only consider two kinds of target sparsity: the local sparsity constraint (LSC) or global sparsity constraint (GSC), which means that the target sparsity is imposed on each layer or the whole network, respectively. 
We consider LSC/GSC for LeNet-5, GSC for LeNet-FC and LSC for VGG-16. We conjecture the performance can be improved if we set the target sparsities layer by layer. 

We only consider global sparsity constrain for LeNet-FC, since the difference between local sparsity constrain and global sparsity constraint of a small network is marginal.  We only consider local sparsity constrain for VGG-16, since for large target sparsity (larger than $0.95$) under global sparsity constraint, some layers of a deep network will be almost totally pruned (with the sparsity almost equal to 1), leading to an unacceptable model. It's too tricky to set varying target sparsities layer by layer for a deep network, which deserves deeper investigation on the weights' importance judgment across layers. 


{\bf Comparison}. To investigate the individual contributions of drop away and drop back, we firstly compare Drop Pruning with: (a) Traditional Pruning, replacing drop away and drop back by deleting all the unimportant weights in Algorithm \ref{alg1} ~\cite{zhu2017prune}, namely setting $p_{\rm away}=1$ and $p_{\rm back}=0$; and (b) Drop away Pruning, the pruning process with only drop away, setting $p_{\rm back}=0$. 
We also compare Drop Pruning to some existing state-of-the-art approaches from two related aspects: weights pruning approaches, including 
LWC~\cite{han2015learning}, DNS~\cite{dns}, 
L-OBS~\cite{dong2017learning}, SWS~\cite{ullrich2017soft}, SNIP~\cite{lee2018snip}, MMP~\cite{zeng2018mlprune}, DTL~\cite{lee2018deeptwist}, DTR~\cite{yeh2018deep}; 
Bayesian training approaches, including GD~\cite{srinivas2016generalized}, SBP~\cite{neklyudov2017structured}, VIB~\cite{dai2018compressing}, DBB~\cite{lee2018adaptive}. Here we ignore the comparison with concrete dropout~\cite{gal2017concrete} and targeted dropout~\cite{Adian2018Targeted} since no corresponding benchmark results are found in their papers. All the results of these comparison approaches directly come from their papers. 
We ignore the comparison between the training and pruning cost for obtaining a sparse model, 
since apparently, our approach takes much time by the requirement of several trails. 
We focus on the on-line size and accuracy of the pruned model. 

{\bf Experimental setup}. To obtain the baseline high accuracy model, we train $18/256$ epochs with an initial learning rate of $0.1/0.1$ and a batch size of $100/128$ for LeNet-5 (LeNet-FC) or VGG-16. For VGG-16, we have the weight decay $3\times 10^{-4}$. 
The pruning periods for LeNet-5/LeNet-FC and VGG-16 are $10$ and $90$ epochs, respectively. We fine-tune the pruned model by $9/90$ epochs for LeNet-5 (LeNet-FC) or VGG-16, respectively. 
By a simple grid search, we let $\xi_1 = 0.9, \xi_2 = 0.08$ for LeNet-5 and LeNet-FC and $\xi_1 = 0.9, \xi_2 = 0.1$ for VGG-16. We conjecture the performance can be improved if we fine-tune these hyper-parameters. 


\begin{table*}[tb]
	\caption{The test errors ($\%$) against varying target sparsities (with corresponding compression ratios) of Traditional Pruning (TP), Drop away Pruning (DaP) and Drop Pruning (DP) for LeNet-5 (LeNet-FC) on MNIST and VGG-16 on CIFAR10 under the local or global sparsity constraint (LSC or GSC). We report the [best, mean, $\pm$std] of the results under $40$ trials for LeNet-5 (LeNet-FC) and $10$ trials for VGG-16. The test errors of baseline models for LeNet-5, LeNet-FC and VGG-16 are $0.74\%$, $2.14\%$ and $7.43\%$ respectively. 
		We report the average results ([best, mean]) under small sparsities ($0.5$-$0.9$) and large sparsities ($0.91$-$0.95$). 
		The best results of fixed target sparsity are in bold. }
	\centering
	\begin{small}
		\scalebox{1}[1]{
			\begin{tabular}{p{30pt}p{10pt}p{62pt}p{62pt}p{62pt}p{62pt}p{62pt}p{40pt}}
				\toprule
				\toprule
				\multicolumn{2}{c}{Sparsity} & \multicolumn{1}{c}{$0.50$}  &\multicolumn{1}{c}{$0.60$} & \multicolumn{1}{c}{$0.70$} & \multicolumn{1}{c}{$0.80$} & \multicolumn{1}{c}{$0.90$} & \multirow{2}{*}{\tabincell{c}{Average}}    \\
				\multicolumn{2}{c}{CR} & \multicolumn{1}{c}{$2\times$}  & \multicolumn{1}{c}{$2.5\times$} & \multicolumn{1}{c}{ $3.3\times$} & \multicolumn{1}{c}{$5\times$}  &\multicolumn{1}{c}{$10\times$}  & \\
				\midrule
				{\multirow{3}{*}{\tabincell{c}{\text{LeNet-5}\\LSC}}}  
				&TP & $[0.64, 0.79, \pm0.08]$&$[\bm{0.54},  0.76,  \pm0.12]$&$[ 0.59, 0.84, \pm0.10]$&$[ 0.71,  0.84, \pm0.07]$&$[0.76, 0.83, \pm0.06]$ & $[0.65, 0.81]$\\
				&\textbf{DaP} & $[0.65, 0.77, \pm0.08]$&$[ 0.66,  0.83,  \pm0.12]$&$[\bm{0.59}, 0.72,  \pm0.11]$&$[ 0.65, 0.85,  \pm0.24]$&$[ 0.75, 0.85, \pm0.09]$ & $[0.66, 0.77]$\\
				&\textbf{DP} & $[\bm{0.58}, 0.79,  \pm0.10]$&$[ 0.61, 0.77, \pm0.10]$&$[ 0.62, 0.79,  \pm0.09]$&$[\bm{0.62}, 0.87, \pm 0.20]$&$[\bm{0.69}, 0.89,  \pm 0.11]$ & $[\bm{0.62}, 0.83]$ \\
				
				\midrule
				{\multirow{3}{*}{\tabincell{c}{\text{LeNet-5}\\GSC}}}  
				&TP & $[ 0.64, 0.68, \pm0.03]$&$[ 0.65, 0.70, \pm0.03]$&$[ 0.65, 0.71, \pm0.03]$&$[ 0.66, 0.72, \pm0.03]$&$[  0.70, 0.74, \pm0.03]$ &$[ 0.66, 0.71]$\\
				&\textbf{DaP} & $[ 0.64, 0.68, \pm0.03]$&$[ 0.65, 0.69, \pm0.02]$&$[ 0.63, 0.68, \pm 0.03]$&$[ 0.61,  0.66,\pm 0.03]$&$[\bm{0.58}, 0.67, \pm0.05]$ &$[ \bm{0.62}, 0.67]$\\
				&\textbf{DP} & $[\bm{0.61}, 0.68, \pm0.03]$&$[\bm{0.65}, 0.68, \pm0.02]$&$[\bm{0.61},  0.68, \pm0.03]$&$[\bm{0.61}, 0.67, \pm0.03]$&$[ 0.62, 0.68, \pm0.04]$ &$[  \bm{0.62}, 0.68]$\\
				
				\midrule
				{\multirow{3}{*}{\tabincell{c}{\text{LeNet-FC}\\GSC}}}  
				&TP & $[ 2.01, 2.07, \pm0.02]$&$[ 1.94, 1.99, \pm0.04]$&$[ 1.95, 1.99,  \pm0.03]$&$[ 2.06, 2.11, \pm0.03]$&$[ 2.13, 2.22, \pm0.06]$ & $[ 2.02, 2.08]$\\
				&\textbf{DaP} & $[ \bm{2.00}, 2.04, \pm0.03]$&$[\bm{1.89}, 1.93,\pm 0.03]$&$[ \bm{1.90}, 1.95, \pm0.03]$&$[  2.00, 2.08,\pm 0.04]$&$[\bm{2.08}, 2.19, \pm0.09]$ &$[ \bm{1.97}, 2.04]$\\
				&\textbf{DP} & $[ 2.06,  2.10, \pm0.02]$&$[ 2.09, 2.17, \pm0.04]$&$[ 2.04,  2.10, \pm0.05]$&$[\bm{1.98}, 2.07,\pm 0.05]$&$[\bm{2.08}, 2.23,   \pm0.12]$ &$[ 2.05, 2.14]$\\
				
				\midrule
				{\multirow{3}{*}{\tabincell{c}{\text{VGG-16}\\LSC}}} 
				&TP & $[ 6.60, 6.84,\pm0.17]$&$[ 6.62,  6.8,  \pm0.13]$&$[ 6.78,  7.00, \pm0.18]$&$[ 7.12, 7.71,   \pm0.43]$&$[ 7.32, 7.44, \pm0.12]$ &$[ 6.89, 7.16]$\\
				&\textbf{DaP}& $[ 6.43, 7.05, \pm0.87]$&$[ 6.37, 6.59,  \pm0.14]$&$[ 6.54, 6.77, \pm0.19]$&$[\bm{6.57}, 6.71, \pm0.09]$&$[ \bm{6.80}, 7.22, \pm0.21]$ &$[ 6.54, 6.87]$\\
				&\textbf{DP} & $[\bm{6.23}, 6.59, \pm0.34]$&$[\bm{6.12}, 6.59, \pm0.17]$&$[\bm{6.53}, 6.96, \pm0.40]$&$[ 6.58, 6.72,\pm 0.11]$&$[ 6.99, 7.26, \pm0.15]$ &$[ \bm{6.49}, 6.82]$\\
				
				\toprule
				\toprule
				
				\multicolumn{2}{c}{Sparsity} & \multicolumn{1}{c}{$0.91$} & \multicolumn{1}{c}{$0.92$}& \multicolumn{1}{c}{$0.93$}& \multicolumn{1}{c}{$0.94$} &  \multicolumn{1}{c}{$0.95$}& \multirow{2}{*}{\tabincell{c}{Average}} \\
				\multicolumn{2}{c}{CR} & \multicolumn{1}{c}{$11.1\times$} &\multicolumn{1}{c}{$12.5\times$} &\multicolumn{1}{c}{$14.3\times$} &\multicolumn{1}{c}{$16.7\times$} & \multicolumn{1}{c}{$20\times$} & \\
				\midrule
				{\multirow{3}{*}{\tabincell{c}{\text{LeNet-5}\\LSC}}}  
				&TP &$[ 0.74, 0.86, \pm0.09]$&$[0.74, 0.91, \pm0.09]$&$[ 0.77, 0.95,  \pm0.13]$&$[ 0.87,  1.09,  \pm0.21]$&$[0.85, 1.03,  \pm0.11]$ & $[ 0.79, 0.97]$\\
				&\textbf{DaP} &$[ \bm{0.70}, 0.98,  \pm0.15]$&$[ 0.83,  1.02,  \pm0.13]$&$[ 0.79,  1.04,  \pm0.26]$&$[ 0.79, 0.87, \pm0.07]$&$[ 0.86,  1.09,  \pm0.25]$ & $[ 0.79, 0.97]$\\
				&\textbf{DP} & $[ 0.73, 0.87,\pm 0.08]$&$[\bm{0.69}, 0.89,  \pm0.11]$&$[\bm{0.73}, 0.89, \pm0.08]$&$[\bm{0.75}, 0.95,\pm 0.10]$&$[\bm{0.85},   1.00,  \pm0.09]$ & $[\bm{0.75}, 0.92]$ \\
				
				\midrule
				{\multirow{3}{*}{\tabincell{c}{\text{LeNet-5}\\GSC}}}  
				&TP & $[ 0.66, 0.75, \pm0.04]$&$[  0.70, 0.76, \pm0.03]$&$[ 0.69, 0.74,  \pm0.03]$&$[ 0.72, 0.76, \pm0.04]$&$[ 0.68, 0.77, \pm0.04]$ &$[ 0.69, 0.76]$\\
				&\textbf{DaP} &$[ 0.62, 0.67, \pm0.04]$&$[\bm{0.57}, 0.68, \pm0.05]$&$[\bm{0.64}, 0.70,\pm 0.03]$&$[\bm{0.64}, 0.73,\pm 0.05]$&$[ 0.66, 0.72, \pm0.04]$ &$[ \bm{0.63}, 0.69]$\\
				&\textbf{DP} & $[ \bm{0.60}, 0.67, \pm0.04]$&$[ 0.61, 0.69,\pm 0.03]$&$[ 0.65, 0.71, \pm0.03]$&$[ 0.67, 0.72,\pm 0.05]$&$[\bm{0.64}, 0.72, \pm0.05]$ &$[ \bm{0.63},   0.70]$\\
				
				\midrule
				{\multirow{3}{*}{\tabincell{c}{\text{LeNet-FC}\\GSC}}}  
				&TP &$[ 2.15, 2.24, \pm0.06]$&$[ 2.15, 2.25, \pm0.07]$&$[ 2.23, 2.31, \pm0.05]$&$[ 2.24, 2.36, \pm0.06]$&$[ 2.27, 2.41, \pm0.07]$ &$[ 2.21, 2.32]$\\
				&\textbf{DaP} & $[ 2.12, 2.19, \pm0.05]$&$[ 2.13, 2.21, \pm0.05]$&$[ 2.17, 2.28, \pm0.08]$&$[ 2.17, 2.31, \pm0.08]$&$[ 2.23, 2.34,  \pm0.06]$ &$[ 2.16, 2.26]$\\
				&\textbf{DP} &$[\bm{2.05}, 2.17, \pm0.10]$&$[ \bm{2.01}, 2.14,  \pm0.10]$&$[\bm{2.05}, 2.16,  \pm0.16]$&$[\bm{2.02}, 2.18, \pm 0.13]$&$[\bm{2.01}, 2.17, \pm0.07]$ &$[ \bm{2.03}, 2.16]$\\
				
				\midrule
				{\multirow{3}{*}{\tabincell{c}{\text{VGG-16}\\LSC}}}
				&TP &$[ 7.42, 7.55, \pm0.10] $&$[ 7.41, 7.56, \pm0.13]$&$[ 7.61, 7.88,  \pm0.33]$&$[ 7.7, 7.91,  \pm0.24]$&$[ 7.97, 8.02, \pm0.05]$ &$[ 7.62, 7.78]$\\
				&\textbf{DaP}&$[\bm{6.84}, 7.36,  \pm0.24] $&$[ 6.96, 7.32, \pm0.20]$&$[ 7.35, 7.59, \pm0.14]$&$[ 7.46, 7.96,   \pm0.45]$&$[ 7.87, 8.03, \pm0.12]$ &$[7.30, 7.65]$\\
				&\textbf{DP} & $[ 6.93, 7.29, \pm0.17]$&$[\bm{6.95}, 7.42, \pm0.24]$&$[\bm{7.21}, 7.62, \pm0.37]$&$[\bm{7.42}, 7.76, \pm0.30]$&$[\bm{7.76}, 8.04, \pm0.18]$ &$[\bm{7.25}, 7.63]$\\
				\bottomrule
			\end{tabular}
		}
	\end{small}
	\label{tab1}
\end{table*}

\begin{figure*}[!htb]
	\begin{center}
		\begin{tabular}{c}
			\includegraphics[width=0.45\textwidth]{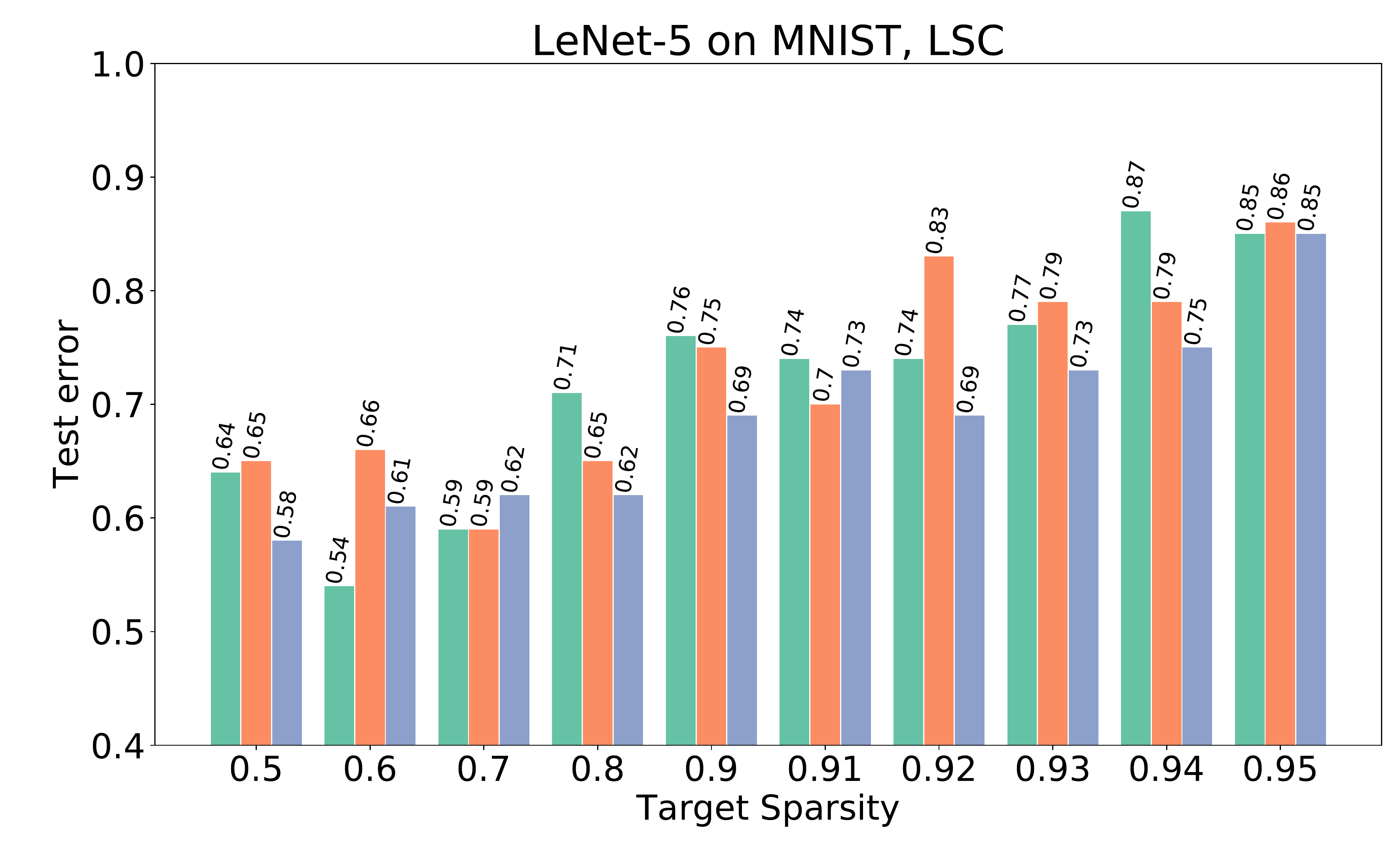}
			\includegraphics[width=0.45\textwidth]{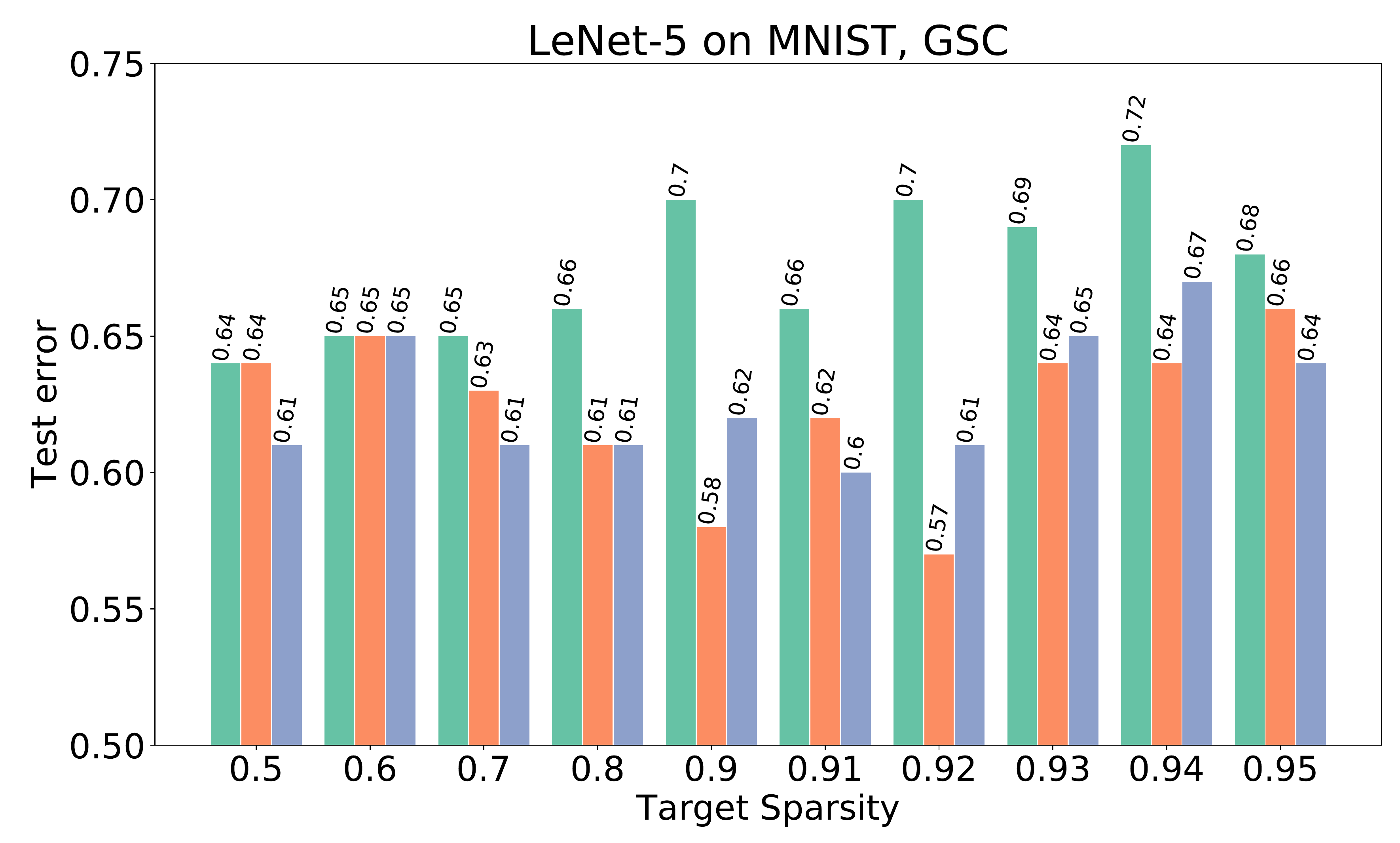}\\
			\includegraphics[width=0.45\textwidth]{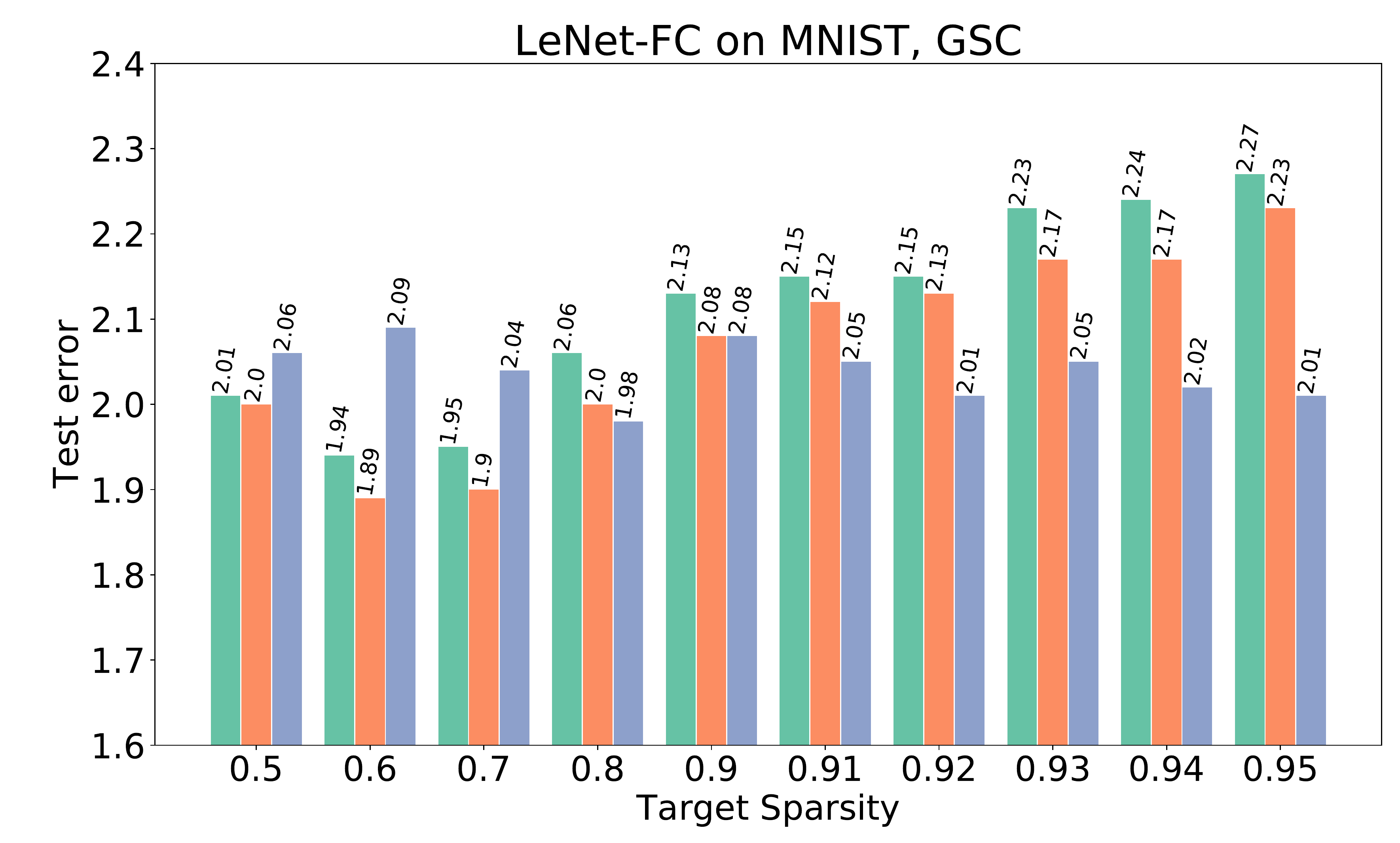}
			\includegraphics[width=0.45\textwidth]{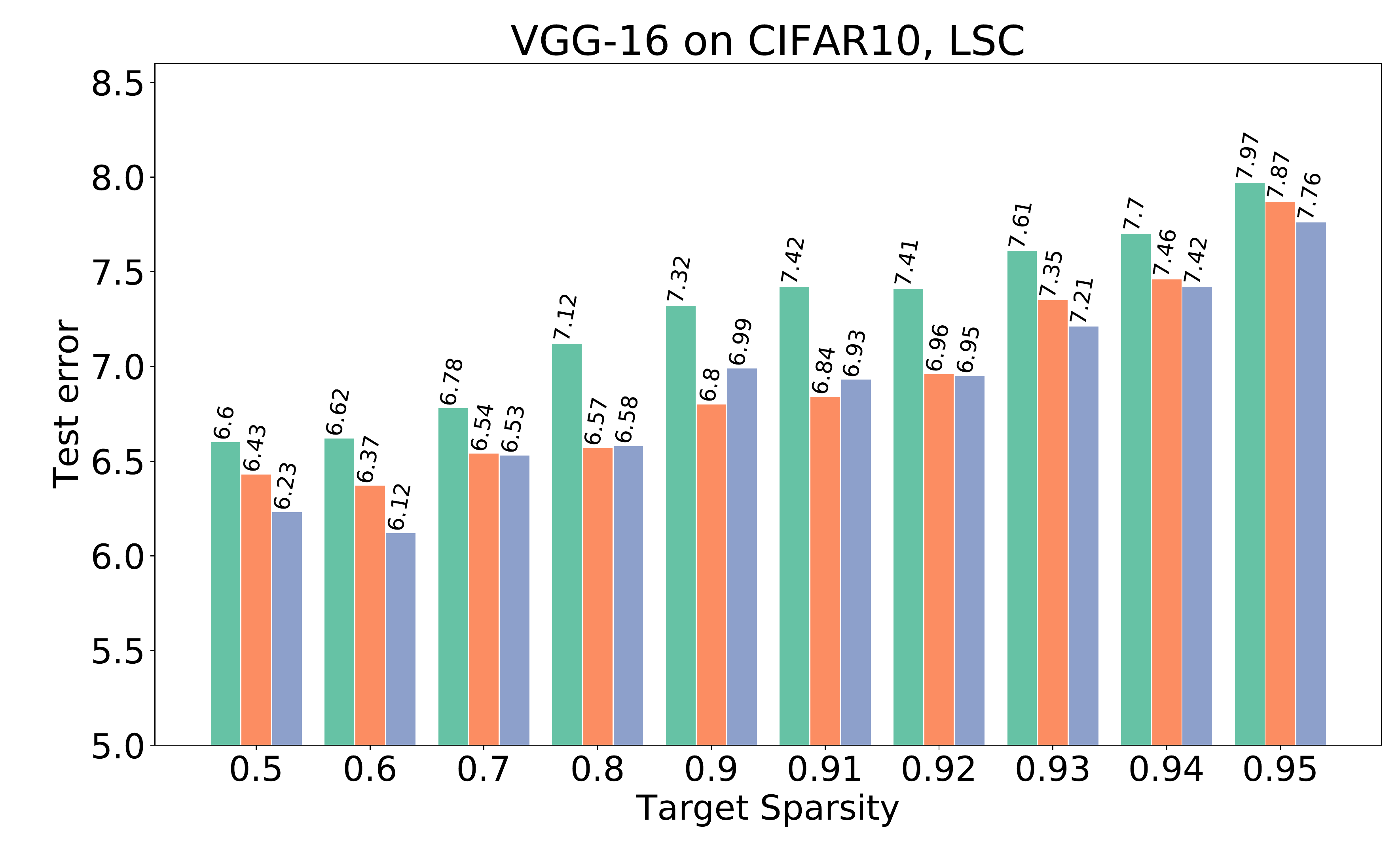}
		\end{tabular}
	\end{center}
	\caption{The test errors of Traditional Pruning (green), Drop away Pruning (orange) and Drop Pruning (blue) against varying pruning tasks. We report the best ones of the results under $10$ trials for VGG-16 and $40$ trials for LeNet-5 and LeNet-FC. The test errors of baseline models for LeNet-5, LeNet-FC and VGG-16 are $0.74\%$, $2.14\%$ and $7.43\%$ respectively. Our approach outperforms the Traditional Pruning at most of the pruning tasks. Here LSC and GSC represents the local sparsity constraint and global sparsity constraint respectively. }\label{fig0}
\end{figure*}

\begin{table*}[!htb]
	\caption{Comparison with related approaches on test error, model size and compression ratio. We report our results by three parts, the one with no (or a litter) accuracy loss than baseline model, the one with best accuracy and the one with best compression ratio (compared with other approaches). These results demonstrate the competitive behavior of our approach. }
	\centering
	\begin{small}
	\scalebox{1}[1]{
		\begin{tabular}{p{24pt}c|ccc|ccc|ccc}
			\toprule
			\toprule
			\multirow{2}{*}{Criterion}  & \multirow{2}{*}{Methods} &  \multicolumn{3}{c|}{LeNet-5 on MNIST}  &   \multicolumn{3}{c|}{LeNet-FC on MNIST} & \multicolumn{3}{c}{VGG-16 on CIFAR10} \\
			& & Test error ($\%$) & Size (K) & CR & Test error ($\%$)  & Size (K) & CR & Test error ($\%$)  & Size (M) & CR \\
			\midrule
			{\multirow{8}{*}{\tabincell{c}{\text{Weights}\\Pruning}}} 
			&  LWC~\cite{han2015learning} &$0.80\rightarrow0.74$ & $10.7$ & $40$ & $1.64\rightarrow1.58$ & $6.8$ & $39$ & --- & --- & ---  \\
			&  DNS~\cite{dns} &$0.91\rightarrow0.91$ & $4.0$ & $108$ & $2.28\rightarrow1.99$ & $4.8$ & $56$ & --- & --- & ---  \\
			&  L-OBS~\cite{dong2017learning} &$1.27\rightarrow1.27$ & $30.1$ & $14.3$ & $1.76\rightarrow1.82$ & $18.6$ & $14.3$ & --- & --- & --- \\
			&  SNIP~\cite{lee2018snip} &$0.80\rightarrow0.80$ & $21.6$ & $20$ & $1.60\rightarrow1.60$ & $13.3$ & $20$ & $6.76\rightarrow7.09$ & $6.5$ & $20$\\
			&  MMP~\cite{zeng2018mlprune} &$0.89\rightarrow0.50$ & $3.7$ & $117$ & $1.86\rightarrow1.94$ & $3.5$ & $77$ & --- & --- & ---  \\
			&  DTL~\cite{lee2018deeptwist} &$0.80\rightarrow0.90$ & $2.2$ & $200$ & $1.60\rightarrow1.90$ & $4.3$ & $62.5$ & --- & --- & ---   \\
			&  DTR~\cite{yeh2018deep} &$0.80\rightarrow0.95$ & $1.7$ & $260$ & --- & --- & ---  & $7.10\rightarrow7.20$ &$2.3$ & $57.1$\\
			&  SWS~\cite{ullrich2017soft} &$0.88\rightarrow0.97$ & $2.7$ & $162$ & $1.89\rightarrow1.94$ & $4.2$ & $64$ & --- & --- & --- \\
			\midrule
			{\multirow{4}{*}{\tabincell{c}{\text{Bayesian}\\Training}}} 
			& GD~\cite{srinivas2016generalized} &$0.67\pm0.02$ & $178$ & $2.42$ & --- & --- & --- & $6.87\pm0.03$ & $26.5$ & $4.90$\\
			& SBP~\cite{neklyudov2017structured} &$0.70\pm0.02$ & $27.8$ & $15.46$ & --- & --- & --- & $7.53\pm0.06$ & $10.5$ & $12.44$\\
			& VIB~\cite{dai2018compressing} &$0.66\pm0.04$ & $37.3$ & $11.43$ & --- & --- & --- & $7.01\pm0.14$ & $13.3$ & $9.75$\\
			& DBB~\cite{lee2018adaptive} &$0.59\pm0.02$ & $35.7$ & $12.08$ & --- & --- & --- & $6.56\pm0.04$ & $15.1$ & $8.59$\\
			\midrule
			{\multirow{3}{*}{\tabincell{c}{\textbf{Drop}\\\textbf{Pruning}}}} 
			& No loss  &$0.74\rightarrow0.64$ & $21.5$ & $20$ & $2.14\rightarrow2.01$ & $13.3$ & $20$ & $7.43\rightarrow7.42$ & $7.8$ & $16.7$\\
			& Best Ac.  &$0.74\rightarrow{\bf 0.58}$ & $215$ & $2$ & $2.14\rightarrow{\bf 1.98}$ & $53.2$ & $5$ & $7.43\rightarrow{\bf 6.12}$ & $52$ & $2.5$\\
			& Best CR  &$0.74\rightarrow0.84$ & $1.8$ & ${\bf 245}$ & $2.14\rightarrow2.19$ & $5.7$ & ${\bf 47}$ & $7.43\rightarrow7.76$ & $6.5$ & ${\bf 20}$\\
			\bottomrule
		\end{tabular}
	}
	\end{small}
	\label{tab2}
\end{table*}

\subsection{LeNet on MNIST}	

We firstly train a baseline model preparing for pruning. The test errors of LeNet-5 and LeNet-FC are $0.74\%$ and $2.14\%$ respectively. 
Starting with the same baseline model, we evaluate the performances of Traditional Pruning, Drop away Pruning and Drop Pruning against varying target sparsities, 
as shown in Table \ref{tab1} with the [best, mean, $\pm$std] of the results from several trails and in Figure \ref{fig0} with the best ones. The overall comparison results demonstrate that our approach performs better and drop away, drop back in Drop Pruning are indeed both significant to achieve a better pruned model, which may because these two drop strategies can help the pruning process escape from the local minimum of $\mathcal{L}_0$. 

Similar with Traditional Pruning, we can also improve the generalization ability of the pruned model under small target sparsities. Under $2\times$ compression ratio, the test error of the pruned model by Drop Pruning for LeNet-5 is only $0.58\%$ which is much less than the one of baseline model. As we discussed in Section 2, that may because our approach would also help the model to escape the local minimum of $\mathcal{L}_D$, i.e. prevent overfitting, similar to dropout~\cite{srivastava2014dropout} or DSD~\cite{han2016dsd}. 
Note for LeNet-FC, the pruned model of Drop Pruning still achieves less test error than baseline model (from $2.14\%$ to $2.01\%$) even under $20\times$ compression ratio. 
We can also achieve $16.7\times$ compression ratio with a litter accuracy loss for LeNet-5 with LSC. 
In addition, we also see the improvement of GSC on LSC of LeNet-5. 
That may because most of the weights in LeNet-5 are located in the fully connected layers. 
Thus using GSC will help to prune more weights in fully connected layers with no accuracy loss. 

We also report the mean and standard deviation of the results under several trails for each pruning task. 
If we compare the average performance (mean value), our approach still has its potential (especially for large target sparsity) even though we always use the best one on-line. 
Besides, our approach roughly has a larger standard deviation than Traditional Pruning. This can be understood intuitively, as our approach introduces more probabilities (from drop away and drop back), while the Traditional Pruning only involves probabilities from SGD. 

Table \ref{tab2} reports the comparisons between our approach with related state-of-the-art approaches and demonstrate the competitive behavior of our approach: comparing with most the other pruning approaches, we can achieve either less test error or high compression rate. We remark that most of the related weights pruning approaches need specifying the target sparsity for each layer, 
while our results were instead evaluated under just local or global sparsity constraint. 


\subsection{VGG-16 on CIFAR10}	
Like LeNet, we firstly train a baseline model with a test error of $7.34\%$. 
Comparing Traditional Pruning, our approach can obtain the best test accuracy under varying target sparsities, 
as also shown in Table \ref{tab1} and Figure \ref{fig0}. 
Under $2\times$ compression ratio, we can also increase the generalization ability of the pruned model, i.e. decreasing the test error to $6.12\%$. 
Furthermore, we can achieve $16.7\times$ compression ratio with a litter accuracy loss. 
We can also obtain the competitive performance with related approaches as shown in Table \ref{tab2}, 
especially compared with Bayesian approaches.


\section{Conclusion}
We have proposed a stochastic model pruning approach, named Drop Pruning, which introduces the stochastic strategy into the pruning. 
Our approach contains two key steps in each pruning step of the gradual pruning process: drop away and drop back, 
which both help the pruning process obtain a better pruned model than Traditional Pruning. 
That may result from that the drop away can reduce the influence of the locally judging weights' importance by its magnitude, 
while the drop back ensures that the pruned weights still have a chance to come back. 
The spirit of drop back also comes from an optimization algorithm with memory mechanism for integer optimization. 
Similar with other pruning approaches, we can also improve the generalization ability of the pruned model under small target sparsity. 
Our approach also has competitive performance than the related weights pruning and Bayesian training approaches. 
This research is in its early stage. There are several aspects that deserve deeper investigation: 
\begin{itemize}
	\item exploit the performance of Drop Pruning with specifying target sparsity for each layer; 
	\item stochastically prune kernels in CNNs; 
	\item extend the idea of drop away and drop back to quantization, which can also be formulated as an integer optimization problem; 
	\item directly extend Drop Pruning to the area of $\mathcal{L}_0$ optimization with nonconvex penalty. 
\end{itemize}

\section*{Acknowledgements}

This work was supported in part by the Innovation Foundation of Qian Xuesen Laboratory of Space Technology, and in part by Beijing Nova Program of Science and Technology under Grant Z191100001119129. 

\bibliographystyle{ieee_fullname}
\bibliography{egbib}

\end{document}